\documentclass[preprint,12pt]{elsarticle}

\usepackage{amssymb}
\usepackage{amsmath}
\usepackage[english]{babel}
\usepackage{CJKutf8}
\usepackage{orcidlink}
\usepackage{hyperref}
\usepackage{url}     
\usepackage{amsfonts}       
\usepackage{nicefrac}     
\usepackage{microtype}     
\usepackage{graphicx}
\usepackage{wrapfig}
\usepackage{epsfig}
\usepackage{amsmath}
\usepackage{amssymb}
\usepackage{subfigure}
\usepackage{marvosym}
\usepackage{color}
\usepackage{threeparttable}
\usepackage[ruled,vlined,linesnumbered]{algorithm2e}
\usepackage{algpseudocode}
\usepackage{setspace}
\usepackage{amsthm}
\usepackage{verbatim}
\usepackage{rotating}  
\usepackage{ulem}
\usepackage{siunitx}

\usepackage{booktabs}
\usepackage{tabularx}
\usepackage{multirow}
\usepackage{array}
\usepackage{makecell}
\usepackage{xcolor}
\usepackage{colortbl}

\definecolor{Green}{RGB}{0,128,80}
\definecolor{lightgray}{gray}{0.9}              
\definecolor{lightblue}{RGB}{220,230,241}       
\definecolor{lightgreen}{RGB}{220,241,220}      

\begin{document}

\begin{CJK*}{UTF8}{gbsn}

\begin{frontmatter}

\title{Dimensional Balance Improves Large Scale Spatiotemporal Prediction Performance}

\author[label1,label3]{Jing Chen\raisebox{4pt}{\resizebox{!}{0.5em}{\orcidlink{0000-0003-3127-8462}}}}
\ead{cj@hdu.edu.cn}
\author[label1]{Shixiang Pan}
\ead{psx@hdu.edu.cn}
\author[label1]{Yujie Fan\raisebox{4pt}{\resizebox{!}{0.5em}{\orcidlink{0009-0001-9996-1519}}}}
\ead{fanyujie@hdu.edu.cn}
\author[label1]{Haocheng Ye}
\ead{yehaocheng@hdu.edu.cn}
\author[label1,label3]{Haitao Xu\raisebox{4pt}{\resizebox{!}{0.5em}{\orcidlink{0000-0002-8387-3776}}}}
\ead{xuhaitao@hdu.edu.cn}
\author[label2]{Wenqiang Xu\raisebox{4pt}{\resizebox{!}{0.5em}{\orcidlink{0000-0003-3378-6719}}}\corref{cor1}}
\ead{wenqiangxu@cjlu.edu.cn}

\cortext[cor1]{Corresponding author}

\affiliation[label1]{
    organization={School of Computer Science and Technology, Hangzhou Dianzi University},
    city={Hangzhou},
    postcode={310018},
    country={China}
}

\affiliation[label2]{
    organization={College of Economics, China Jiliang University},
    city={Hangzhou},
    postcode={314423},
    country={China}
}

\affiliation[label3]{
    organization={Key Laboratory of New Industrial Internet Control Technology},
    city={Hangzhou},
    postcode={310018},
    country={China}
}

\begin{abstract}
Accurate spatiotemporal pattern analysis is critical in fields such as urban traffic, meteorology, and public health monitoring. However, existing methods face performance bottlenecks, typically yielding only incremental gains and often exhibiting limited cross-domain transferability. We analyze this bottleneck through spatial and temporal entropy measures, which are used as diagnostic indicators of spatiotemporal complexity mismatch rather than as guarantees that entropy alignment alone yields better forecasting. Empirically, larger mismatch is often accompanied by higher prediction uncertainty, especially under a fixed model-capacity budget. Guided by this diagnostic, we propose a scalable, adaptive framework that harmonizes spatial and temporal feature representations. Spatial dimensionality is compressed via low-rank matrix embedding to preserve essential structure, while an extended temporal horizon captures long-range dependencies and mitigates cumulative errors arising from temporal heterogeneity. Extensive experiments on urban traffic, meteorological, and epidemic datasets demonstrate substantial accuracy gains and broad applicability across the evaluated domains, suggesting that the framework is promising for a wide range of spatiotemporal tasks beyond the current study. The code is available on GitHub at \url{https://github.com/ST-Balance/ST-Balance}.
\end{abstract}


\end{frontmatter}

\section{Introduction}
Accurate spatiotemporal prediction supports numerous critical applications\cite{jin2024survey}, including urban traffic management\cite{chen2025bilinear, chen2026spatiotemporal}, meteorological forecasting\cite{bi2023accurate, Corrformer2023}, and public health monitoring\cite{kraemer2025artificial, Epi-Cola-GNN2023}. These domains require methods capable of simultaneously capturing complex spatial patterns and temporal dynamics to provide precise and forward looking insights. However, as datasets become increasingly complex, many traditional methods fail to sustain consistent predictive performance across different scales. This challenge underscores the importance of balancing the relative scales of spatial and temporal feature dimensions within predictive models.

Recent advances in spatiotemporal prediction, particularly frameworks that combine Graph Neural Networks (GNNs) with Recurrent Neural Networks\cite{DCRNN2018, GWNet2019} or Transformers\cite{STAEformer2023, DTRformer2025} models, have shown significant progress by integrating node neighborhood features with historical temporal information. These approaches have achieved impressive results in diverse real-world scenarios. Yet, as spatial networks grow, effectively integrating large-scale spatial information with limited temporal information becomes increasingly difficult, often leading to a greater prediction errors. This phenomenon highlights the need for strategies that ensure robust performance on large, heterogeneous datasets.
\begin{figure}[htbp]
    \begin{center}
    \includegraphics[width=0.8\linewidth]{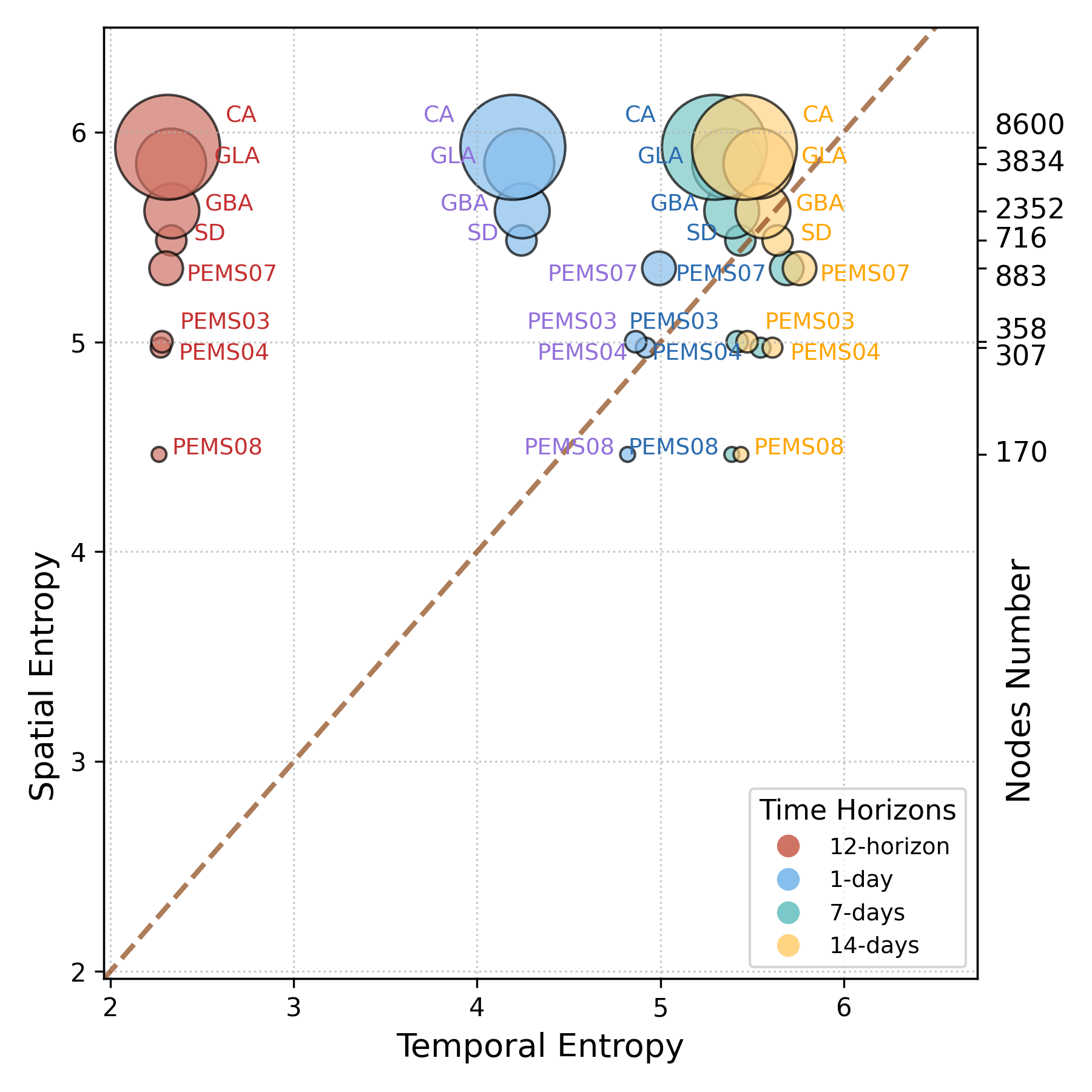}
    \end{center}
    \caption{\textbf{Scatter plots of spatial entropy vs. temporal entropy for four distinct time horizons (12-horizon, 1-day, 7-day, and 14-day).} Marker size is proportional to the number of nodes in each network, and a diagonal reference line indicates where spatial and temporal entropies are equal. Points further away from the diagonal indicate greater mismatch between spatial and temporal entropy, which is empirically associated with increased modeling difficulty and higher predictive uncertainty. Larger spatial scales correspond to higher spatial entropy, requiring richer temporal context to reduce mismatch under limited model capacity.}
    \label{fig:Imbalance_Ratio} 
\end{figure}

To explore the underlying mechanism, we selected eight traffic flow datasets\cite{PEMS, largeST2024} spanning a range of network scales and examined their spatiotemporal structure under multiple temporal horizons (daily, weekly, biweekly, and a commonly used 12-horizon rolling window). We quantified both spatial and temporal entropy (see Supplementary, 1 for details) and plotted each dataset’s temporal entropy on the x-axis and spatial entropy on the y-axis (Figure~\ref{fig:Imbalance_Ratio}). We observed that as the number of nodes increases, overall spatial complexity and uncertainty rise, leading to higher estimated spatial entropy. In contrast, temporal entropy under commonly used short windows is constrained by predictive range, sampling frequency, and input length. Moreover, for a fixed network scale, the spatial entropy estimated from the fixed prior graph remains relatively stable across temporal horizons, while temporal entropy increases as the look-back window grows. 

Rather than treating entropy balance as a mathematical guarantee of accuracy, we use it as an interpretable diagnostic of complexity mismatch under limited capacity: when the intrinsic spatial complexity (large \(H_S\)) is paired with a shallow temporal context (small \(H_T\)), or when excessively long windows make temporal variability dominate, models tend to exhibit higher error and uncertainty. Importantly, spatial dimensionality reduction in our framework does not change the underlying graph; instead, it compresses the effective degrees of freedom of spatial representations presented to the predictor, making the mismatch easier to handle with finite model capacity.

In addition to factors like sensor distribution, correlation structures, and data quality, our findings show that the balance between spatial and temporal scales also critically affects predictive accuracy. When spatial complexity increases without adequate temporal balance or constraint, models often operate in a higher-uncertainty regime and tend to exhibit larger errors. Conversely, larger network scales can offer richer representations of underlying phenomena, which may enhance prediction performance if the information is effectively captured. Mitigating such high entropy complexity may involve decomposing spatial complexity or augmenting temporal data to provide sufficient constraints and learning signals. Nevertheless, effectively utilizing this broader spatiotemporal context remains challenging, since computational overhead often restricts input windows to a 12 horizon. Even long-window pre-trained models (e.g. STEP\cite{STEP2022} and STD-MAE\cite{STD-MAE2024}) must rely on truncation methods to feed data into spatiotemporal architectures such as GWNet\cite{GWNet2019}, where memory and runtime costs are considerable. These limitations underscore the importance of rigorous model design, particularly when reconciling broad spatial domains with constrained historical data windows.

In this work, the entropy scores are primarily used to guide design decisions and to motivate why certain configurations become difficult at scale. We do not claim that aligning \(H_S\) and \(H_T\) alone is sufficient to guarantee better forecasting, because the optimal operating point depends on model capacity, regularization, and the data-generating process. Our goal is to provide a practical and measurable principle for scalable model design, supported by extensive empirical evidence.
To address this issue, we propose a novel framework designed to mitigate the effects of spatiotemporal imbalance by reducing the disparity between spatial and temporal entropy. First, to effectively handle large-scale node networks, we reduce spatial dimensionality through low-rank matrix embedding and node aggregation, while preserving prior graph knowledge. This approach significantly constrains computational complexity without sacrificing essential structural information. Second, building upon this spatial dimensionality reduction, we extend the temporal dimension, enhancing the model's capacity to capture long-term dependencies and reduce cumulative errors arising from extensive spatial heterogeneity. The integration of these strategies ensures that our framework remains scalable and robust as network complexity and size continue to increase.

We conducted extensive experiments on diverse, large-scale datasets encompassing traffic flow, meteorological, and epidemic scenarios. The results show that our framework not only improves prediction accuracy but also enhances adaptability across different data scales and task requirements. By resolving this core bottleneck in spatiotemporal prediction, our work provides both theoretical and practical foundations for subsequent scalable, dimensionally balanced modeling strategies, offering new insights into the prediction of complex spatiotemporal phenomena.

\section{Related Work}
\subsection{Graph Neural Network-Based Spatiotemporal Models}
Early deep models combined CNNs for spatial encoding with RNN or TCN modules for temporal dynamics. Research then shifted to spatiotemporal graph neural networks that exploit sensor topology to model spatial dependence and long horizons. Existing STGNNs can be broadly divided into static graph and dynamic graph methods. These two families differ in how spatial relations are obtained, how much prior structure is used, and how well the model scales when the number of nodes increases.

\subsubsection{Static Graph Neural Networks}
Static graph methods use a fixed adjacency and focus on effective coupling of spatial and temporal operators. STGCN\cite{STGCN2020} couples spectral graph convolution with temporal gated convolution. DCRNN\cite{DCRNN2018} models directed diffusion with sequence to sequence GRUs. Graph WaveNet\cite{GWNet2019} augments a distance graph with a learned adaptive adjacency. AGCRN\cite{AGCRN2020} learns node embeddings and relations without a given graph. Recent work further targets robustness and scalability. STWave\cite{STWave} introduces wavelet spectral modules for multi scale signals. STD-MAE\cite{STD-MAE2024} uses spatial temporal masked pretraining. BigST\cite{BigST2024} redesigns operators for near linear complexity on large graphs.

The main advantage of static graph methods is that physical or precomputed topology provides a stable spatial inductive bias, which improves local propagation when the graph is reliable. These methods are also easier to train than fully dynamic graph models because the spatial structure is not repeatedly inferred. However, a fixed graph can miss context-dependent relations, and a globally learned graph may still retain redundant or noisy connections when the network becomes large. The cost of message passing also grows with graph size, so large prior graphs can dominate the learning signal from short temporal windows.

\subsubsection{Dynamic Graph Neural Networks}
Dynamic graph methods infer time-varying relations. STTN\cite{STTN2020} uses attention to learn directed dependencies that change with context. STGODE\cite{STGODE} casts evolution as a continuous time process. DGCRN\cite{DGCRN2023} generates step wise adjacencies with hyper networks. D$^2$STGNN\cite{D2STGNN} decouples components and updates graphs accordingly. FlashST\cite{FlashST2024} adapts pre-trained models across cities through spatiotemporal prompts.

These designs are useful when correlations change with traffic conditions, weather regimes, or epidemic spread. They improve flexibility by adapting the graph to the current temporal context. Their limitation is that relation generation, attention, or hypernetwork modules increase memory and latency. In large networks, dynamic relation estimation is harder to regularize and can become unstable, especially when the temporal input is short and the number of nodes is large. Thus, dynamic graphs relax the fixed-topology assumption, but their computational overhead and optimization difficulty can become bottlenecks at scale.

\subsection{Graph-Free Models}
Graph-free models remove graph convolutions and predefined adjacency and rely on embeddings and attention to learn spatial relations from data. STID\cite{STID2022} attaches spatial and temporal identity embeddings to each series and uses a lightweight MLP to capture location specificity and periodicity without any graph. ST-Norm\cite{STNorm2021} addresses distribution heterogeneity with separate spatial and temporal normalization that stabilizes optimization across nodes and time. Building on the identity based idea, STAEformer\cite{STAEformer2023} adopts a pure Transformer with a serial design. A temporal self attention encoder models each node's dynamics first, then a spatial self attention encoder models cross node interactions. DTRformer\cite{DTRformer2025} follows the same core idea but uses a parallel design. Temporal and spatial Transformer encoders run concurrently and their outputs are fused by cross spatial and temporal attention.

These models are simple to train and scale well when prior graphs are unavailable or unreliable. They can avoid the computational burden of explicit message passing and remain competitive on standard traffic benchmarks. Their limitation is that spatial dependence must be learned from scratch. When the physical topology is informative, discarding the prior graph may reduce robustness, particularly for large road networks or county-level epidemic systems with meaningful connectivity. Spatial attention can also become expensive when node counts are high, while identity embeddings may overfit if many nodes have sparse or uneven observations.

\subsection{Research Gaps and Positioning}
The literature progresses from CNN and RNN baselines to static and dynamic STGNNs and then to graph-free architectures. Each family improves a different part of the spatiotemporal forecasting pipeline, yet several gaps remain for large-scale prediction.
\begin{itemize}
\item Most methods improve the spatial operator or temporal encoder separately, but they do not explicitly regulate the effective spatial capacity relative to the available temporal context when the node number grows.
\item Prior graph methods preserve topology, but they often keep high-dimensional graph representations, which increases memory cost and can make spatial information dominate temporal learning.
\item Long-history information is useful for periodic and slowly changing patterns, but many spatiotemporal models still rely on short input windows because combining long sequences with large graphs is expensive.
\item Graph-free models scale well, but they may discard useful physical priors that become important in large systems with reliable connectivity.
\end{itemize}

These gaps motivate ST-Balance. Rather than replacing existing predictors with a new type of backbone, ST-Balance rebalances the effective inputs seen by the predictor. Low-rank spatial embedding compresses prior structure while preserving local connectivity, and temporal window expansion provides longer historical constraints. This design targets scalable forecasting when spatial complexity and temporal context are mismatched.

\section{Overview}
\begin{figure*}[htbp]
	\begin{center}
		\includegraphics[width=\linewidth]{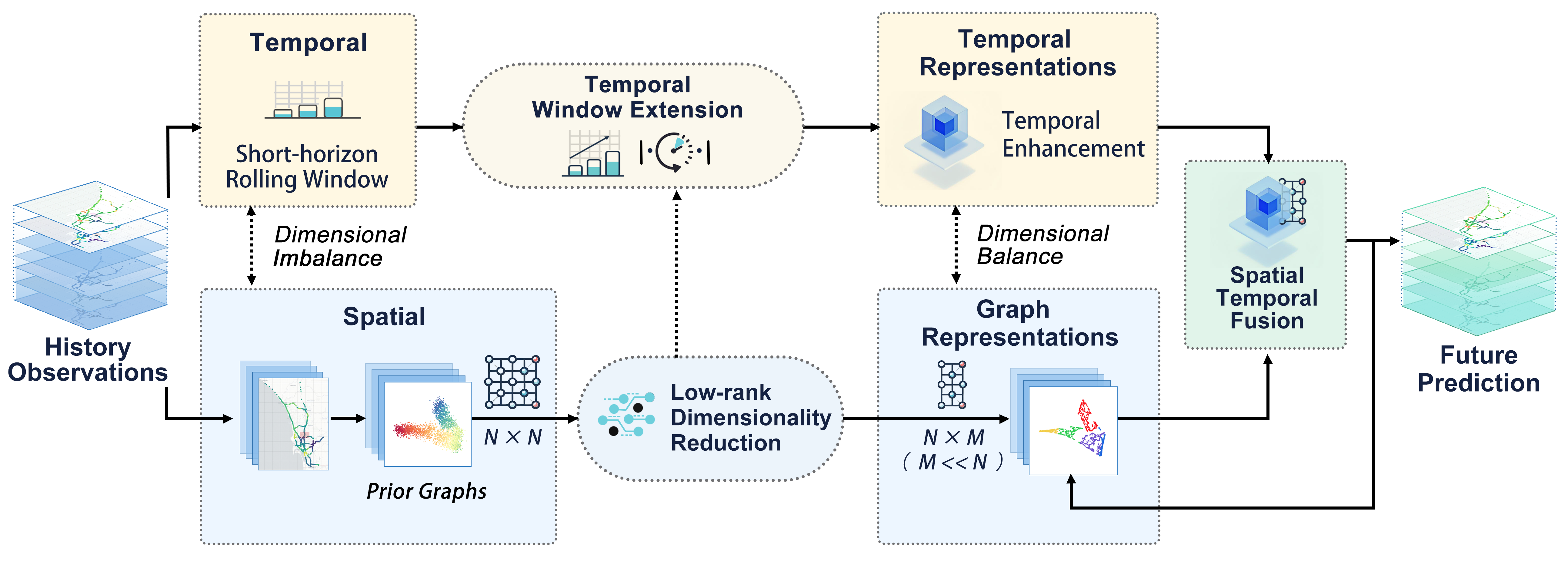}
	\end{center}
	\caption{\textbf{Framework overview integrating spatial dimensionality reduction, temporal window expansion, and fusion.} The left side indicates the initial dimensional imbalance between the short-horizon temporal input and the prior graphs, while the right side indicates the balanced representations used for fusion.}
	\label{fig:Our_Model}
\end{figure*}

The workflow depicted in Figure~\ref{fig:Our_Model} outlines the transformation of historical observations into accurate future predictions by addressing critical dimensional imbalances between spatial and temporal features.

A significant challenge emerges from the disproportionate scale between the spatial dimension, represented by large sparse $N\times N$ adjacency matrices, and the relatively limited temporal scope. Processing these spatially extensive yet temporally shallow data sets directly is computationally intensive and risks neglecting critical local connectivity details.

The spatial pathway addresses the challenge of dimensional overload by embedding the extensive $N\times N$ spatial graphs into compact lower-rank $N\times M$ matrices, with $M\ll N$. This strategic dimensionality reduction preserves vital local connectivity patterns while eliminating redundancy and aligns spatial complexity with the expanded temporal scope.

Spatial dimensionality reduction decreases model complexity and frees computational capacity to accommodate an expanded temporal window. To enhance temporal resolution, the temporal pathway extends beyond the conventional short-range rolling window and incorporating broader historical intervals. A Temporal Enhancement module then integrates recent fluctuations with long-term patterns, generating refined temporal representations that more effectively capture essential temporal dynamics.

After spatial dimensionality reduction and temporal window expansion, the model achieves a balanced representation of spatial and temporal features. These spatially compressed graph representations and enhanced temporal features are then integrated within the Spatial–Temporal Fusion module. With balanced dimensionality, this module efficiently captures complex interdependencies between spatial connectivity and temporal evolution. The resulting fused features enable robust and accurate predictions, concluding the workflow.

\section{Method}
\subsection{Spatial Dimensionality Reduction Retaining Local Features}
\subsubsection{Limitations of Standard Dimensionality Reduction} 
In large-scale networks, the prior graph \( \mathbf{A} \in \mathbb{R}^{N \times N} \) is typically sparse (\(n \ll N^2\)), which makes naive dense factorization or dense matrix operations impractical at scale.

The sparsity of the graph leads to inefficiencies in the application of traditional dimensionality reduction methods, such as PCA\cite{PCA} and UMAP\cite{UMAP}. These methods perform poorly on sparse prior graphs because the decomposed graph becomes dense, which increases storage and computation costs.
Methods relying on distance metrics, such as Node2Vec\cite{Node2Vec} and HOPE\cite{HOPE}, also perform inadequately in sparse spaces, as the definition of distance becomes ambiguous in such contexts. Moreover, global methods fail to capture critical local connectivity patterns in network data, which are important for accurate representation.

\subsubsection{Spatial Dimensionality Reduction Using Low-Rank Matrices}
Let $\mathbf{A}\in\mathbb{R}^{N\times N}$ denote the prior graph adjacency, with edge set $\mathcal{E}$.
Our goal is to obtain a compact spatial representation $\mathbf{H}\in\mathbb{R}^{N\times M}$ with $M\ll N$,
which preserves local connectivity while keeping both memory and runtime scalable.

We use the graph Laplacian $\mathbf{L}=\mathbf{D}-\mathbf{A}$ to motivate low-rank spatial representations.
Since $\mathbf{L}$ is symmetric positive semidefinite, a truncated eigendecomposition
$\mathbf{L}=\mathbf{U}\boldsymbol{\Lambda}\mathbf{U}^\top$ admits a rank-$M$ approximation
$\mathbf{L}\approx \mathbf{U}_M\boldsymbol{\Lambda}_M\mathbf{U}_M^\top
= \mathbf{H}_{\mathrm{spec}}\mathbf{H}_{\mathrm{spec}}^\top$,
where $\mathbf{H}_{\mathrm{spec}}=\mathbf{U}_M\boldsymbol{\Lambda}_M^{1/2}$.
This spectral form provides a principled motivation for representing spatial structure using a compact matrix
$\mathbf{H}\in\mathbb{R}^{N\times M}$.

In implementation, to avoid explicit eigendecomposition on large sparse graphs and to prevent any dense $N\times N$ operations during forecasting, we parameterize $\mathbf{H}$ with a lightweight neural projection and optionally refine it using a sparse reconstruction objective defined only on the support of the prior graph.

Concretely, we maintain a learnable node embedding table $\mathbf{E}\in\mathbb{R}^{N\times M}$ and compute
\begin{equation}
\mathbf{H} = \mathbf{E} + \mathrm{FC}_2\bigl(\sigma(\mathrm{FC}_1(\mathbf{E}))\bigr).
\end{equation}
To encourage $\mathbf{H}$ to respect the prior topology, we optionally minimize a reconstruction loss on edges:
\begin{equation}
\mathcal{L}_{\mathrm{rec}} =
\sum_{(u,v)\in\mathcal{E}}\bigl(A_{uv}-\mathbf{h}_u^\top\mathbf{h}_v\bigr)^2
+\beta\sum_{(u,v)\in\mathcal{N}}\bigl(\mathbf{h}_u^\top\mathbf{h}_v\bigr)^2,
\end{equation}
where $\mathcal{N}$ is a negative sample set and $\mathbf{h}_u$ denotes the $u$-th row of $\mathbf{H}$.
This refinement is computed on sparse pairs and thus scales with $|\mathcal{E}|+|\mathcal{N}|$ rather than $N^2$.
Importantly, the spatial module is used for representation compression, while cross-node interactions are modeled later in the fusion module using reduced features.

The rank $M$ controls spatial capacity: too small $M$ may underfit and lose key local structures, while too large
$M$ reduces compression benefits. In practice, we select $M$ by validation and observe that moderate ranks preserve
community/cluster structures and improve downstream forecasting accuracy.

\subsection{Temporal Window Expansion}
To address the issues of capacity imbalance and error accumulation in spatiotemporal forecasting, we propose a temporal window expansion method, which consists of two key components.
We denote the multivariate spatiotemporal input as $\mathbf{X}\in\mathbb{R}^{N\times T\times F}$, where
$N$ is the number of nodes, $T$ is the look-back length, and $F$ is the feature dimension.
For readability, the following temporal equations omit the node dimension and are applied independently to each
node with shared parameters, unless stated otherwise. Long-range history is first projected and then aligned to
the short-scale length $T_{\text{short}}$ so that long- and short-scale features can be concatenated along the
feature/channel dimension.
\begin{figure}[htbp]
	\centering
	\includegraphics[width=0.6\linewidth]{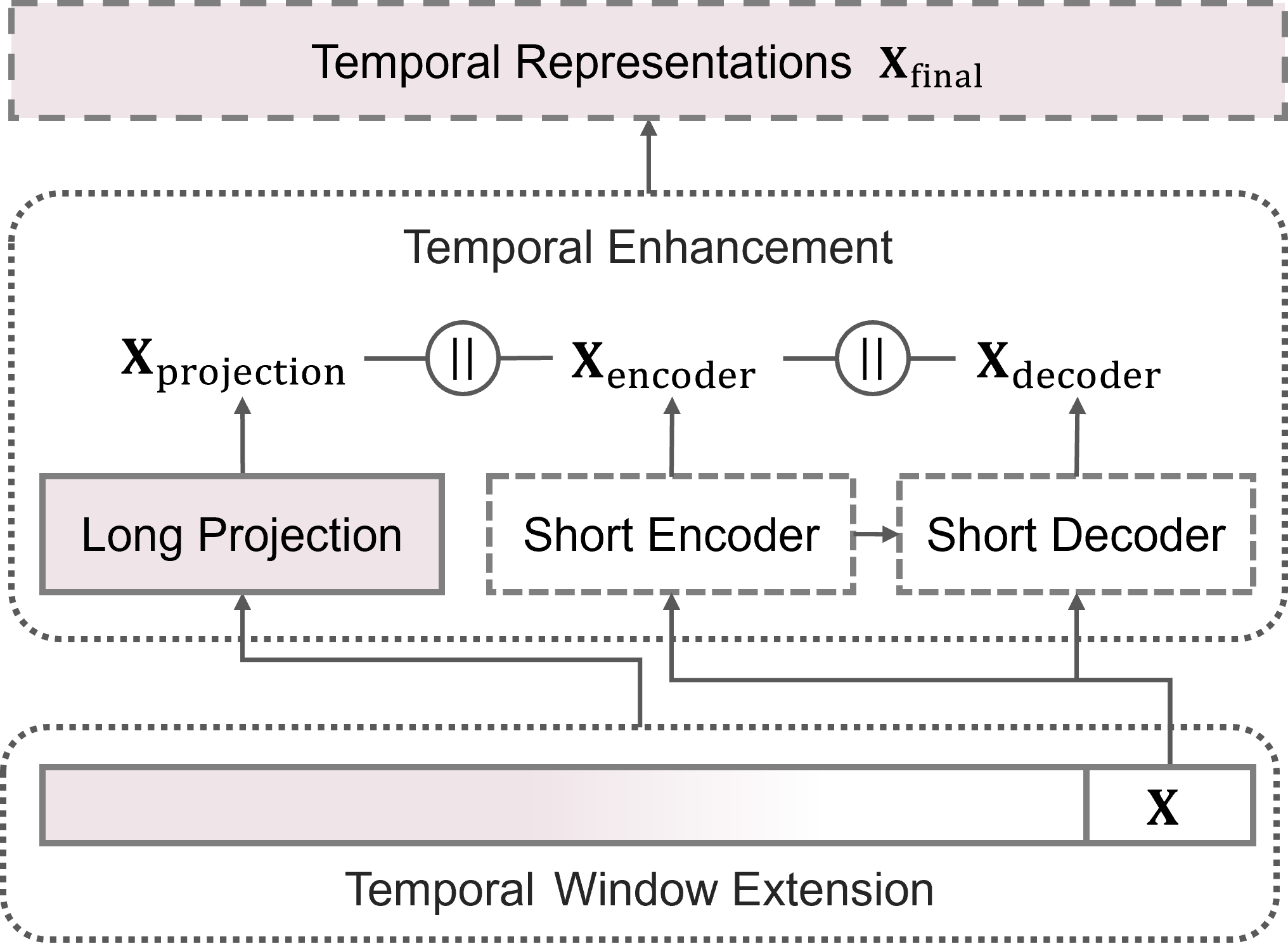}
	\caption{\textbf{Temporal enhancement module.} The short-term window shown here is one instantiation of the Temporal Enhancement strategy.}
	\label{fig:method-T}
\end{figure}

\subsubsection{Temporal Window Extension}
By extending the temporal window, we enhance the model's ability to capture long-term dependencies, thereby increasing temporal capacity. 
For a given node $i$, let the original time-series input be $\mathbf{X}_i\in\mathbb{R}^{T\times F}$, and we extend it to $T'>T$ to capture longer historical information.
The extended features are then transformed by a projection module to \( \mathbf{X}_{\text{proj}} \), which integrates the extended temporal information.

The temporal error after extension is given by:
\begin{equation}
	E_{T,i}^{t'} = g_{\text{temporal}}\left( \left\{ \mathbf{h}_i^\tau \mid \tau \leq t' \right\} \right) - \mathbf{s}_i^{t'}.
\end{equation}
where $\mathbf{s}_i^{t'}$ is the ground-truth value for node $i$ at horizon $t'$, and $g_{\text{temporal}}(\cdot)$ denotes the temporal backbone that maps historical representations to the prediction.

Since the temporal features are richer, the model fits \( g_{\text{temporal}}(\cdot) \) better, thus reducing the expected temporal error:

\begin{equation}
    \mathbb{E}\left[ |E_{T,i}^{t'}| \right] \le \mathbb{E}\left[ |E_{T,i}^t| \right], \quad \text{for } T < T' \le T^\star,
\end{equation}
where \(T^\star\) denotes a data- and capacity-dependent saturation point. The above inequality is a heuristic motivation that holds when the additional historical context provides non-redundant predictive signal and the temporal backbone has sufficient effective capacity. For \(T' > T^\star\), the marginal benefit can diminish or even reverse due to redundancy, noise accumulation, or distribution shift, which is consistent with the non-monotonic behavior observed in our ablation studies.
By increasing the temporal information, the model effectively reduces the contribution of the temporal error component, leading to a lower overall prediction error.

\subsubsection{Temporal Enhancement}
The network structure, as shown in the figure, is designed as a multi-scale temporal feature extraction network to unify temporal representations across both long-time spans and short-time local intervals. The core components of this network include a long-time projection module, a short-time encoder, and a short-time decoder.

The long-time projection module is designed to reduce or initially encode long-span temporal data to obtain long-term features. Its output is expressed as:
\begin{equation}
	\mathbf{X}_{\text{proj}}^{(\text{long})}
	\in 
	\mathbb{R}^{T_{\text{long}}' \times d_{p}},
\end{equation}
where \( T_{\text{long}}' \) is the number of time steps after projection, and \( d_{p} \) is the feature dimension after projection. This process alleviates the dimensional burden for downstream processing while retaining important patterns over the global temporal range. 

The short-time encoder introduces a network structure where multi-head attention (MHA) and residual multi-layer perceptron (RMLP) are alternately stacked, allowing for better capture of high-resolution temporal information within short time segments. Let \( \mathbf{X} \in \mathbb{R}^{T_{\text{short}} \times d_{x}} \) represent the short-time sequence input. The encoder updates layer by layer as follows:

\begin{equation}
	\begin{aligned}
		&\mathbf{U}_{\text{enc}}^{(0)} = \mathbf{X}^{\text{(short)}}, \\[6pt]
		&\mathbf{U}_{\text{enc}}^{(\ell)} = f_{\ell}\bigl(\mathbf{U}_{\text{enc}}^{(\ell-1)}\bigr),
		\quad 
		\ell = 1,\dots,L_{E}, \\[6pt]
		&\mathbf{X}^{\text{(short)}}_{\text{enc}} = \mathbf{U}_{\text{enc}}^{(L_E)},
	\end{aligned}
\end{equation}
where 
\( \mathbf{U}_{\text{enc}}^{(0)} = \mathbf{X}^{\text{(short)}} \) is the initial input to the short-time encoder, which is the short-time sequence itself.
\( f_{\ell}(\cdot) \) represents the composite function at the \( \ell \)-th layer, consisting of multi-head self-attention, residual connections, layer normalization, and RMLP.

After stacking \( L_E \) layers, the encoder outputs \( \mathbf{X}^{\text{(short)}}_{\text{enc}} \), which represents the deep temporal representation at the local time scale.
The short-time decoder is structurally similar to the encoder, also using the stack of multi-head attention and residual multi-layer perceptron, but its initialization and attention interaction differ slightly. As required, we initialize the \textbf{decoder} the same as the \textbf{encoder}:

\begin{equation}
	\begin{aligned}
		\mathbf{V}_{\text{dec}}^{(0)} &= \mathbf{U}_{\text{enc}}^{(0)} \;=\; \mathbf{X}^{\text{(short)}},
	\end{aligned}
\end{equation}
to ensure that the decoder initially receives the same raw short-time sequence input as the encoder. Subsequently, each layer of the decoder can interact with the encoder output \( \mathbf{X}_{\text{encoder}} \) during the multi-head attention phase to better reconstruct or refine the short-time features. This is formalized as:

\begin{equation}
	\begin{aligned}
		&\mathbf{V}_{\text{dec}}^{(\ell)} = g_{\ell}\bigl(\mathbf{V}_{\text{dec}}^{(\ell-1)},\, \mathbf{X}^{\text{(short)}}_{\text{enc}}\bigr),
		\quad
		\ell = 1,\dots,L_{D}, \\[4pt]
		&\mathbf{X}^{\text{(short)}}_{\text{dec}} = \mathbf{V}_{\text{dec}}^{(L_{D})}.
	\end{aligned}
\end{equation}

Similarly, \( g_{\ell}(\cdot) \) includes multi-head attention, residual connections, layer normalization, and MLP as basic units, but its attention mechanism selectively incorporates the encoder output to achieve more flexible feature fusion.
After extracting \( \mathbf{X}_{\text{proj}}^{(\text{long})} \) (long-time features), \( \mathbf{X}^{\text{(short)}}_{\text{enc}} \) (short-time encoded features), and \( \mathbf{X}^{\text{(short)}}_{\text{dec}} \) (short-time decoded features), the network can fuse all three, for example, through concatenation:
\begin{equation}
	\mathbf{X}_{\text{final}}
	\;=\;
	\mathbf{X}_{\text{proj}}^{(\text{long})}
	\;\|\;
	\mathbf{X}^{\text{(short)}}_{\text{enc}}
	\;\|\;
	\mathbf{X}^{\text{(short)}}_{\text{dec}},
\end{equation}
resulting in a multi-scale unified representation that combines long-term global information with short-term local fine-grained details.
Here $\|$ denotes concatenation along the feature/channel dimension after aligning the temporal length to $T_{\text{short}}$.

\subsection{Hierarchical Spatiotemporal Fusion Model}
\begin{figure*}[htbp]
	\centering
	\includegraphics[width=\linewidth]{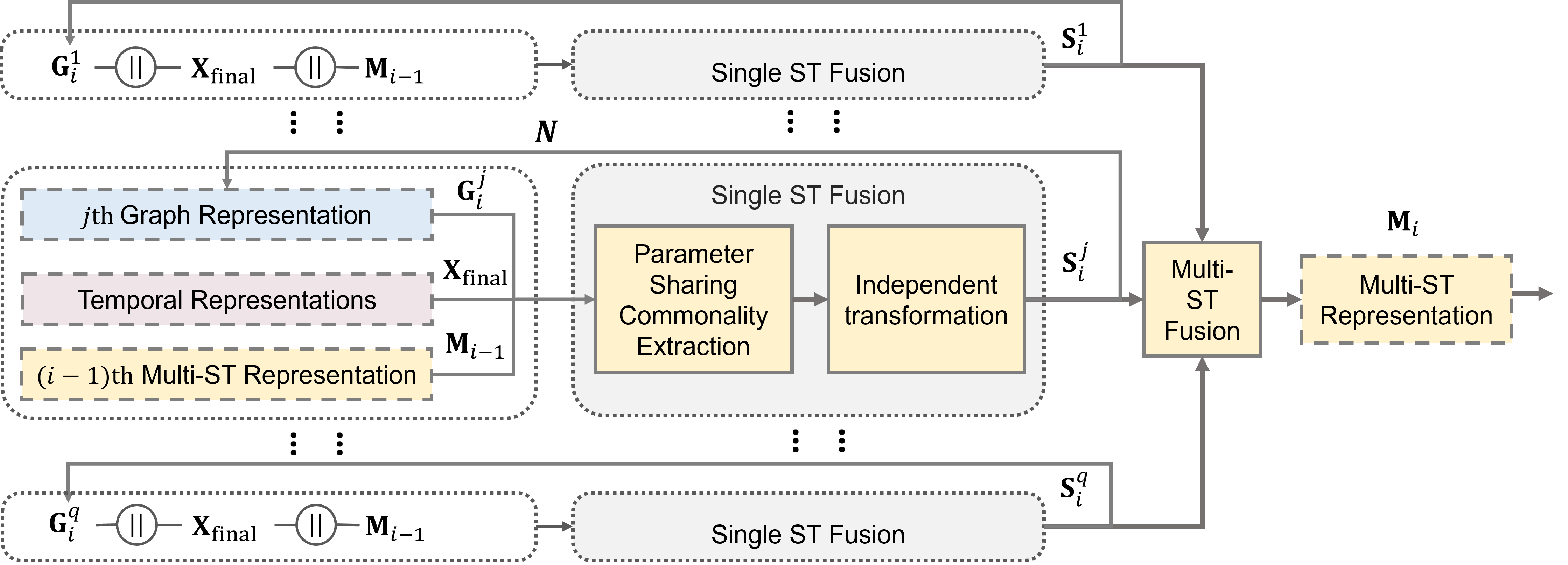}
	\caption{\textbf{Hierarchical spatiotemporal fusion model.}}
	\label{fig:method-F}
\end{figure*}

We fuse temporal features $\mathbf{X}_{\text{final}}$ with dimension-reduced spatial features $\mathbf{H}$ using a shared parameter spatiotemporal fusion module (STFM) stacked for $L$ layers (Figure~\ref{fig:method-F}). Sharing across $J$ prior graphs captures common propagation patterns and avoids the parameter blow-up of multi-graph GNNs, while light graph-specific heads provide specialization.

\subsubsection{Hierarchical Fusion Strategy}
The framework stacks \(L\) layers of STFM, each comprising two stages: single-ST fusion (SF) and multi-ST fusion (MF). We denote the reduced spatial embedding by \(\mathbf{H}\in\mathbb{R}^{N\times M}\). For each prior graph \(j\in\{1,\ldots,J\}\), we initialize graph-specific spatial features by \(\mathbf{G}_0^j = \phi^j(\mathbf{H})\), where \(\phi^j(\cdot)\) is a lightweight (e.g., linear) projection. We set the initial fused state \(\mathbf{M}_0\) to a zero representation with the same channel dimension as \(\mathbf{M}_i\).
In the single fusion stage, the spatial features \(\mathbf{G}_i^j \ (j=1,\ldots,J)\) at layer \(i\) are modeled with fine-grained spatiotemporal interaction modeling to produce an intermediate output \(\mathbf{S}_i^j\). 
Here \(\mathbf{S}_i^j\) denotes the single-graph fused representation at layer \(i\) for prior graph \(j\).
\begin{equation}
	\mathbf{S}_i^j = \mathrm{FC}_i^j\Bigl(\mathrm{SF}_i\bigl(\mathbf{X}_{\text{final}} \  \Vert  \ \mathbf{G}_i^j  \ \Vert  \ \mathbf{M}_{(i-1)}) \Bigr),
\end{equation}
capturing the internal dependencies among features. In the multi-fusion stage, the output for this layer is obtained as \(\mathbf{M}_i\)
\begin{equation}
	\mathbf{M}_i = \mathrm{FC}_i\Bigl(\mathrm{MF}_i\bigl(\mathbf{S}_i^1 \  \Vert \  \cdots \  \Vert \  \mathbf{S}_i^J\bigr)\Bigr).
\end{equation}
By progressively refining information layer by layer, this divide-and-conquer approach avoids the feature overshadowing or noise amplification that can arise in a one-step fusion.

\subsubsection{Intermediate Feature Feedback}
In each layer, the output \(\mathbf{S}_i^j\) updates the spatial features \(\mathbf{G}_{i+1}^j\) through self-attention and nonlinear transformations, dynamically enhancing representational capacity. Here $\mathrm{Self\text{-}Attention}(\cdot)$ is used as a lightweight, node-wise gating mechanism,
i.e., it reweights the feature channels of $\mathbf{G}_i^j$ without mixing information across different nodes,
and therefore does not incur dense $O(N^2)$ token-mixing cost.

\begin{equation}
	\mathbf{G}_{i+1}^j = \mathbf{S}_i^j \odot \sigma\bigl(\mathrm{FC}(\mathrm{Self\text{-}Attention}(\mathbf{G}_i^j))\bigr),
\end{equation}
where \(\odot\) denotes elementwise multiplication and \(\sigma\) is an activation function. 
In our implementation, the self-attention term is used as a lightweight gating mechanism to reweight spatial features, which preserves scalability while still enabling adaptive feature refinement.

The model employs a residual multi-layer perceptron for feature-fusion encoding, where each layer satisfies
\begin{equation}
	(\mathbf{Z})^{(l+1)}
	= \mathrm{FC}_2^l\Bigl(\sigma\bigl(\mathrm{FC}_1^l(\mathbf{Z}^l)\bigr) + \mathbf{Z}^l\Bigr).
\end{equation}
This architecture preserves continuity in feature transmission while its hierarchical stacking progressively improves the accuracy of modeling dominant spatiotemporal dynamics.

\section{Results}
We conducted a series of experiments to evaluate the ST-Balance framework from multiple perspectives. We first assess overall predictive performance compared to state-of-the-art baselines, establishing whether the proposed balance between temporal and spatial feature dimensions indeed boosts spatiotemporal forecasting. We then examine the effectiveness of each component in ST-Balance, followed by tests of robustness, computational efficiency, and applicability across real-world domains. The following subsections detail how each experiment connects to our overarching goal of delivering a more balanced spatiotemporal prediction methodology.

\subsection{Performance and Comparative Analysis}


 \begin{table}[htbp]
\centering
\caption{\textbf{Performance comparison of ST-Balance and baseline models on multi-scale traffic flow datasets.} Showing Mean Absolute Error (MAE, lower is better) across datasets spanning small (PEMS03, PEMS04, PEMS08), medium (PEMS07, LargeST SD), and large (LargeST GBA, GLA, CA) scales. Models are grouped according to graph-type (static, dynamic, or without graph neural networks). Cross-hatched cells indicate computational infeasibility (out-of-memory, OOM). ST-Balance consistently achieves superior accuracy across all scales.}
\label{tab:performance-MAE}

\scriptsize
\setlength{\tabcolsep}{3pt} 
\renewcommand{\arraystretch}{1.1} 

\begin{tabularx}{\linewidth}{@{}l *{8}{>{\centering\arraybackslash}X} @{}}
\toprule
\rowcolor{lightblue}
\multirow{2}{*}{\textbf{Method}} & \multicolumn{3}{c}{\textbf{Small Scale}} & \multicolumn{2}{c}{\textbf{Medium Scale}} & \multicolumn{3}{c}{\textbf{Large Scale}} \\
\cmidrule(lr){2-4} \cmidrule(lr){5-6} \cmidrule(lr){7-9}
\rowcolor{lightblue}
& \textbf{PEMS03} & \textbf{PEMS04} & \textbf{PEMS08} & \textbf{PEMS07} & \textbf{SD} & \textbf{GBA} & \textbf{GLA} & \textbf{CA} \\
\rowcolor{lightblue}
& \textbf{358 nodes} & \textbf{307 nodes} & \textbf{170 nodes} & \textbf{883 nodes} & \textbf{716 nodes} & \textbf{2352 nodes} & \textbf{3834 nodes} & \textbf{8600 nodes} \\
\midrule

\rowcolor{lightgray}
\multicolumn{9}{l}{\textbf{Classical (without Spatial)}} \\
HL & 23.81 & 31.56 & 25.28 & 35.45 & 60.79 & 56.44 & 59.58 & 54.10 \\
LSTM & 16.55 & 22.22 & 16.23 & 23.23 & 26.44 & 27.96 & 28.05 & 26.89 \\
\addlinespace[2pt]

\rowcolor{lightgray}
\multicolumn{9}{l}{\textbf{GNN-based (with Spatial)}} \\
\rowcolor{lightgreen}
\multicolumn{9}{l}{\textit{Static Graph}} \\
DCRNN, 2018 & 15.88 & 20.43 & 15.93 & 21.32 & 21.03 & 23.13 & 23.17 & 21.87 \\
STGCN, 2020 & 15.83 & 19.63 & 15.98 & 21.94 & 19.67 & 23.42 & 22.64 & 21.33 \\
GWNet, 2018 & 14.50 & 18.82 & 14.41 & 20.37 & 17.74 & 20.91 & 21.20 & 21.72 \\
AGCRN, 2020 & 15.30 & 19.25 & 15.39 & 20.68 & 18.09 & 21.01 & 20.25 & OOM \\
STWave, 2023 & 15.18 & 18.53 & 13.96 & 19.65 & 18.22 & 20.81 & 20.96 & 19.69 \\
STD-MAE, 2024 & \textbf{13.80} & 17.80 & 13.44 & 18.65 & OOM & OOM & OOM & OOM \\
BigST, 2024 & 15.88 & 18.16 & 12.94 & 18.41 & 18.80 & 21.95 & 22.08 & 20.32 \\
\addlinespace[2pt]

\rowcolor{lightgreen}
\multicolumn{9}{l}{\textit{Dynamic Graph}} \\
ASTGCN, 2019 & 17.82 & 21.07 & 17.83 & 24.37 & 23.70 & 26.47 & 28.99 & OOM \\
STGOD, 2021 & 16.13 & 19.86 & 15.35 & 20.96 & 19.55 & 21.79 & 21.49 & 20.77 \\
STTN, 2020 & 15.92 & 19.23 & 15.51 & 20.81 & 18.69 & 20.97 & OOM & OOM \\
DSTAGNN, 2022 & 15.57 & 19.30 & 15.67 & 21.42 & 21.82 & 23.82 & 24.13 & OOM \\
D$^2$STGNN, 2022 & 14.34 & 18.26 & 14.38 & 19.91 & 17.85 & 20.71 & OOM & OOM \\
DGCRN, 2023 & 14.47 & 18.74 & 14.29 & 20.14 & 18.02 & 20.91 & OOM & OOM \\
FlashST, 2024 & 15.52 & 18.35 & 14.47 & 20.17 & 18.84 & 21.46 & OOM & OOM \\
\addlinespace[2pt]

\rowcolor{lightgray}
\multicolumn{9}{l}{\textbf{Non-GNN-based (with Spatial)}} \\
\rowcolor{lightgreen}
\multicolumn{9}{l}{\textit{Without Priors Graph}} \\
STNorm, 2021 & 15.24 & 19.13 & 15.40 & 20.54 & 18.23 & 21.55 & 21.82 & 19.91 \\
STID, 2022 & 15.30 & 18.28 & 14.20 & 19.61 & 17.89 & 20.58 & 20.29 & 19.11 \\
STAEformer, 2023 & 14.91 & 18.13 & 13.36 & 19.31 & 17.63 & 21.41 & 20.37 & 19.59 \\
DTRformer, 2025 & 14.50 & 17.54 & 12.59 & 18.01 & OOM & OOM & OOM & OOM \\
\addlinespace[2pt]

\rowcolor{lightgreen}
\multicolumn{9}{l}{\textit{With Priors Graph}} \\
BLSTF, 2025 & 14.05 & 17.93 & 13.49 & 18.87 & 17.33 & 19.79 & 19.52 & 18.46 \\
\textbf{ST-Balance} & 13.87 & \textbf{17.59} & \textbf{12.82} & \textbf{18.29} & \textbf{15.06} & \textbf{16.59} & \textbf{15.78} & \textbf{14.93} \\
\bottomrule
\end{tabularx}
\end{table}

\begin{figure*}[htpb]
	\begin{center}
	\includegraphics[width=\linewidth]{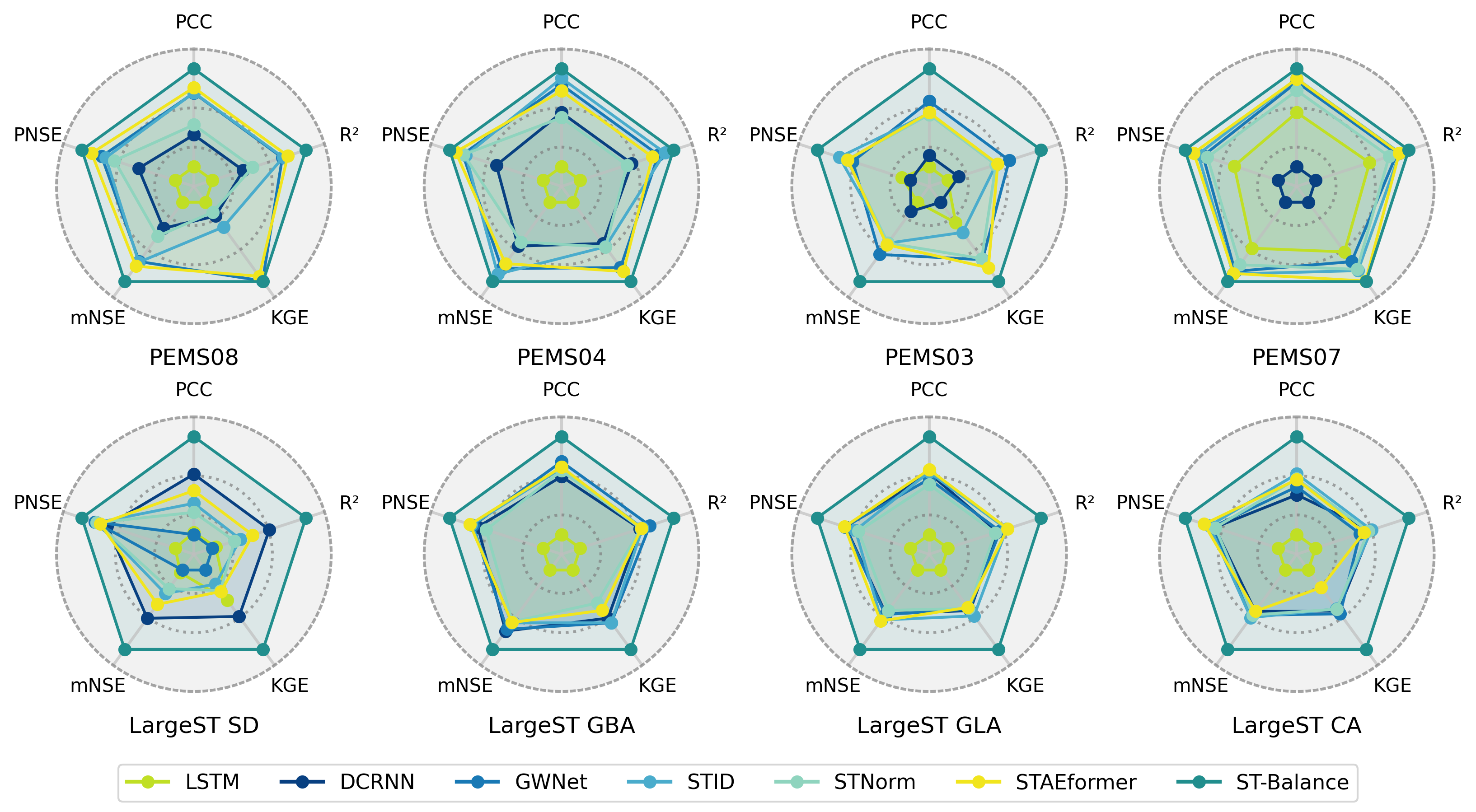}
	\end{center}
	\caption{\textbf{Statistical comparison of multi-indicator performance between ST-Balance and baseline models across diverse traffic datasets.} Radar plots illustrate statistical results for Pearson correlation coefficient (PCC), coefficient of determination (R$^2$), Kling-Gupta efficiency (KGE), modified Nash–Sutcliffe efficiency (mNSE), and percentage of Nash–Sutcliffe efficiency (PNSE). ST-Balance demonstrates superior correlation, interpretability, and dynamic prediction capabilities on datasets spanning small-scale to large-scale complex traffic conditions.}
	\label{fig:static_performance}
\end{figure*}

Table~\ref{tab:performance-MAE} compares mean absolute error (MAE) (Supplementary, 4.B
for error metrics and Supplementary, 5.A.1 
for additional results) between ST-Balance and various state-of-the-art baselines (details in Supplementary, 4.A.1
). ST-Balance consistently achieves lower MAE. Figure~\ref{fig:static_performance} further demonstrates that ST-Balance yields higher correlation, interpretability, and dynamic performance (see Supplementary, 4.B.2 
for statistical indicators and Supplementary, 5.A.3 
for further details) than competing methods across all scales and prediction horizons (Supplementary, 5.A.2 
), including on the highly complex and computationally intensive LargeST CA dataset.

Although GNN-based models generally perform well on smaller and medium-scale datasets, many encounter performance bottlenecks, reflected by Out-of-Memory (OOM) errors, on large-scale datasets such as LargeST GLA and LargeST CA. For instance, dynamic graph models (e.g., STTN\cite{STTN2020}, STGODE\cite{STGODE}, DGCRN\cite{DGCRN2023}, D$^2$STGNN \cite{D2STGNN}, FlashST\cite{FlashST2024}) excel at capturing temporal fluctuations yet require substantial computational resources, limiting their scalability\cite{largeST2024}. Static graph models (e.g., DCRNN\cite{DCRNN2018}, GWNet\cite{GWNet2019}, STGCN\cite{STGCN2020}, AGCRN\cite{AGCRN2020}, STWave\cite{STWave}, BigST\cite{BigST2024}) partially alleviate memory constraints but typically show reduced accuracy relative to non-GNN linear methods (e.g., STID\cite{STID2022}, STNorm\cite{STNorm2021}, STAEformer\cite{STAEformer2023}, DTRformer\cite{DTRformer2025}, BLSTF\cite{chen2025bilinear}) under increased spatial complexity.

These observations align with the premise described in the Introduction: as node counts increase, spatial entropy may surpass the constraining capacity of the temporal dimension, resulting in an imbalance between spatial and temporal feature dimensions. By emphasizing control over spatial complexity and integrating sufficiently rich temporal or external information, ST-Balance remains robust as dataset size and complexity escalate. Its strong performance on LargeST CA and other large datasets highlights the value of a balanced spatiotemporal representation in high entropy environments, where naive graph expansions or purely deep architectures may struggle to sustain both accuracy and efficiency.

\subsection{Effectiveness of ST-Balance’s Core Components}
The strong performance of ST-Balance derives from two core strategies aimed at mitigating imbalances in the spatiotemporal dimensions: (i) spatial dimensionality reduction, and (ii) temporal expansion. A low rank matrix embedding approach is used to reduce spatial dimensionality, thereby curbing node dimensionality. This process reduces model complexity and limits overfitting risks, which is especially important when modeling large scale networks with thousands of nodes. With this computational relief, ST-Balance then expands its temporal receptive field to capture long-term dependencies often neglected by conventional spatiotemporal models constrained by high graph complexity.

\subsubsection{Verification of Spatial Dimensionality Reduction Strategy}
\begin{figure}[htbp]
	\begin{center}
		\includegraphics[width=0.6\linewidth]{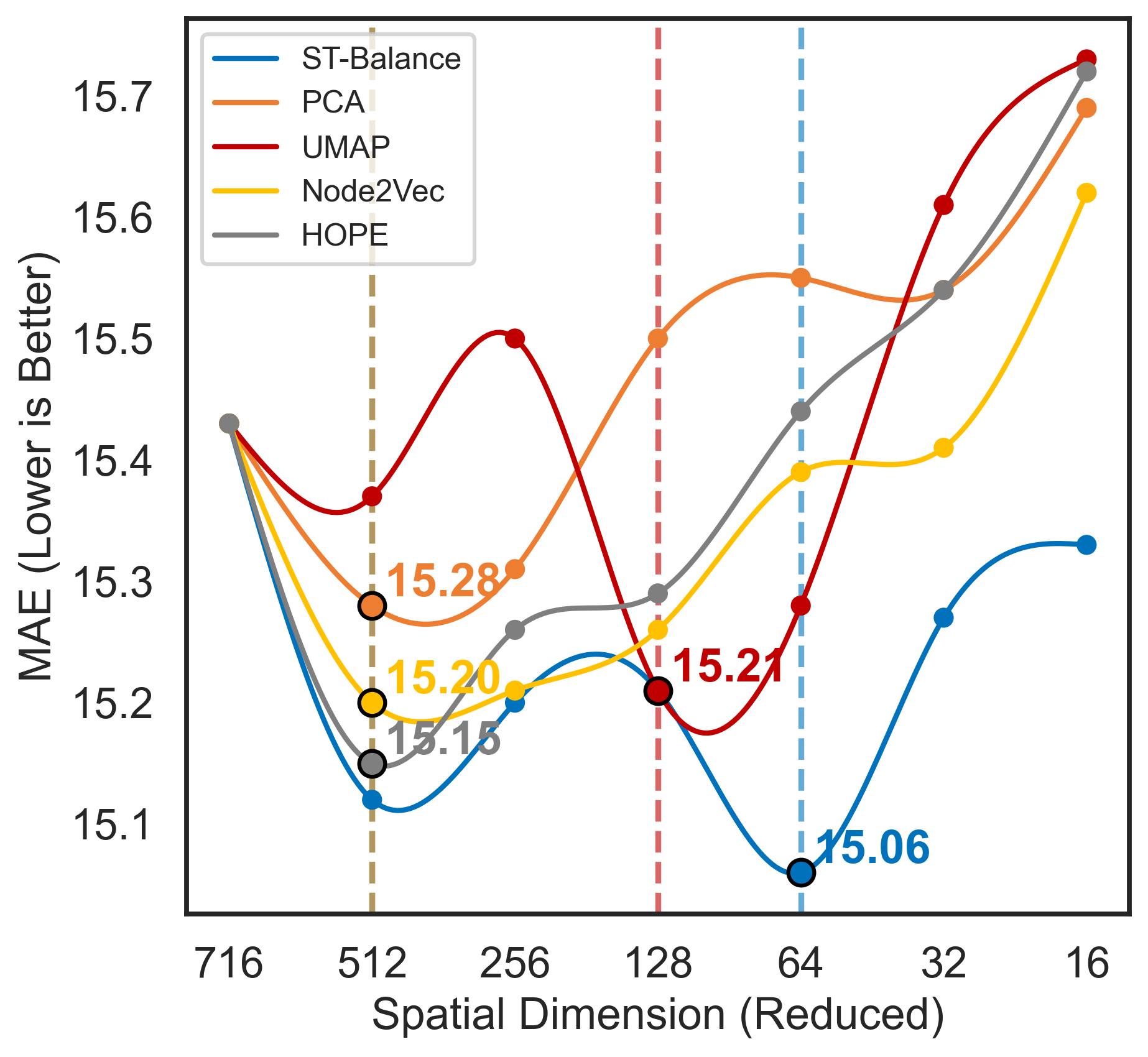}
	\end{center}
	\caption{\textbf{Prediction error under varying spatial dimensionality reduction methods.} Comparison of prediction performance on the LargeST SD dataset when reducing spatial embedding dimension from 716 to 16 using PCA\cite{PCA}, UMAP\cite{UMAP}, Node2Vec\cite{Node2Vec}, HOPE\cite{HOPE}, and our low-rank embedding method. All methods initially benefit when the dimension is reduced from 716 to 512. However, classical linear methods degrade markedly below 512 due to loss of complex structural information, whereas our approach consistently preserves accuracy, demonstrating effective retention of structural features in sparse spatiotemporal data. The ST-Balance curve is non-monotonic because the reduced rank controls both compression and spatial capacity: moderate ranks suppress redundant graph variations, while very small ranks merge distinct traffic patterns.}
	\label{fig:reduce_methods}
\end{figure}

To investigate the potential of spatial dimensionality reduction in enhancing spatiotemporal prediction accuracy, we evaluated several standard dimensionality reduction techniques (Supplementary, 4.A.4 
) against our proposed low-rank embedding approach on the LargeST SD dataset as shown in Figure \ref{fig:reduce_methods}. Our findings indicate that moderate dimensionality reduction effectively mitigates spatial-temporal entropy imbalance. Unlike traditional linear methods, our method maintained prediction accuracy at significantly reduced dimensions, highlighting its suitability for preserving critical nonlinear and structural relationships within large-scale, sparse spatiotemporal networks.

\begin{figure*}[htbp]
	\begin{center}
		\includegraphics[width=\linewidth]{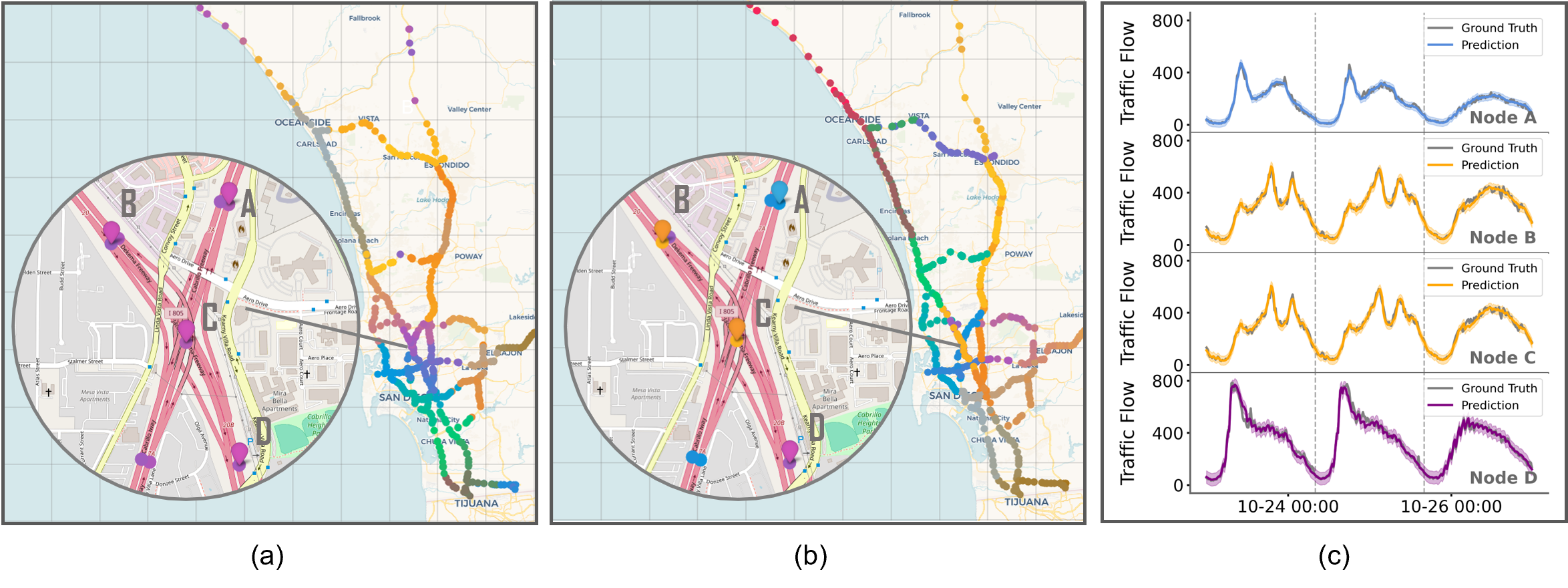}
	\end{center}
	\caption{\textbf{t-SNE views of SD traffic-node embeddings. Colors denote flow patterns.} (a) Original embeddings are scattered. (b) Low-rank dimensionality reduction yields clear, flow-consistent clusters. (c) Zoomed hub: similar nodes (B,C) co-locate; dissimilar nodes (A,D) separate.
    }
	\label{fig:space_visual}
\end{figure*}

The ST-Balance curve in Figure~\ref{fig:reduce_methods} is non-monotonic because the embedding rank controls both compression and spatial capacity. When the dimension is close to the node number, the representation still contains redundant and weakly informative variations from the sparse graph, and the temporal module has less effective capacity to constrain them. Moderate reduction removes part of this redundancy and improves accuracy. When the rank becomes too small, nodes with different traffic patterns are forced into similar embeddings, which loses local structure and raises the error. The small secondary improvement at lower ranks reflects that the SD network contains a limited set of dominant flow patterns, as also shown by the flow-consistent clusters in Figure~\ref{fig:space_visual}. Once the rank is below the diversity of these patterns, the error increases again. Therefore, the spatial dimension is selected by validation rather than minimized.

Figure~\ref{fig:space_visual} shows how spatial dimensionality reduction via low-rank embedding enhances structural clarity within traffic networks. Initially, the node distribution lacks clear organization, complicating identification of coherent flow patterns. However, after applying dimensionality reduction, distinct clusters emerge, notably among nodes belonging to identical road segments or sharing similar flow behaviors. This clustering is particularly evident around transportation hubs, where nodes previously indistinct in the original embedding clearly differentiate into groups with parallel or divergent traffic characteristics. Such structural clarity underscores the capability of low-rank methods to extract meaningful spatial patterns, enabling efficient representation of complex spatiotemporal dynamics.

\begin{figure*}[htbp]
	\begin{center}
		\includegraphics[width=\linewidth]{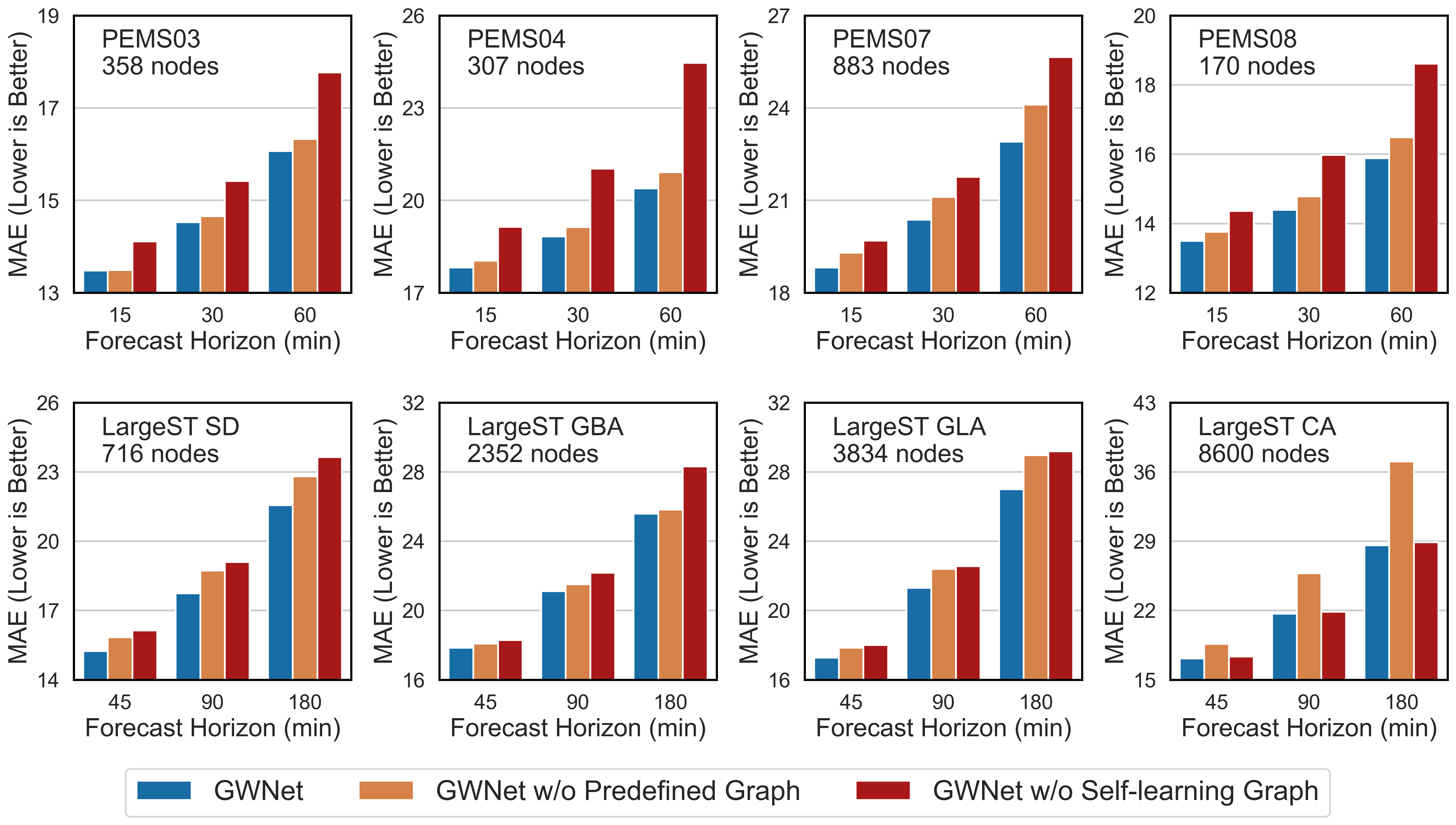}
	\end{center}
	\caption{\textbf{Scale-dependent contributions of graph structures.} Performance (MAE) changes when selectively removing prior or adaptive graphs from GWNet, highlighting the critical role of prior structures in larger networks.}
	\label{fig:prior-adptive}
\end{figure*}

\begin{figure*}[htbp]
	\begin{center}
		\includegraphics[width=\linewidth]{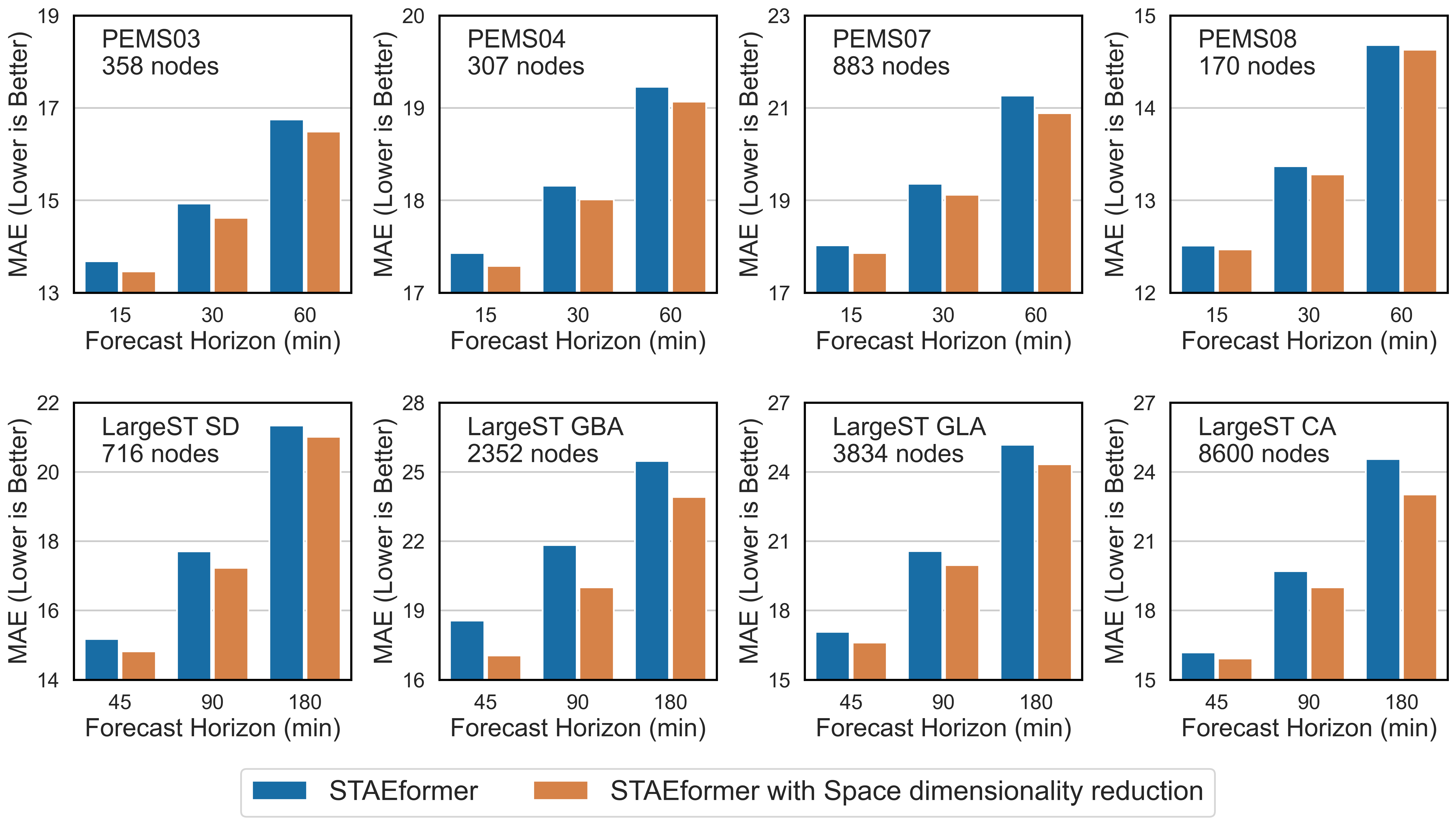}
	\end{center}
	\caption{ \textbf{Scale-dependent benefits of dimensionality reduction.} Performance improvements (MAE) from incorporating low-rank dimensionality reduction into STAEformer, highlighting pronounced gains in larger networks.}
	\label{fig:dimensionality_reduction}
\end{figure*}

The dimensionality reduction capability of ST-Balance emerges from the application of a low-rank embedding that preserves prior graph structures. To elucidate the complementary functions of prior and adaptive graphs, we selectively modified the GWNet model\cite{GWNet2019} by disabling each component. Figure~\ref{fig:prior-adptive} illustrates performance variations across datasets of different scales, demonstrating that smaller and medium-sized networks primarily rely on adaptive graph structures for accuracy, a trend aligned with recent adaptive graph methodologies\cite{STID2022,STNorm2021,STAEformer2023}. Importantly, large-scale datasets (LargeST GBA, GLA, CA) exhibit greater dependence on prior physical graph structures due to their intrinsic topological complexity. Reinforcing this observation, we integrated our low-rank spatial dimensionality reduction method, informed explicitly by prior knowledge, into the leading STAEformer model. As shown in Figure \ref{fig:dimensionality_reduction}, this enhancement yields significant improvements, particularly on large-scale datasets. These findings underscore the necessity of combining adaptive mechanisms with established structural priors, highlighting that achieving spatial-temporal balance is essential for robust predictions in complex, large-scale systems.

\subsubsection{Analysis of the Impact of the Time Window}
\begin{figure*}[htbp]
	\begin{center}
		\includegraphics[width=\linewidth]{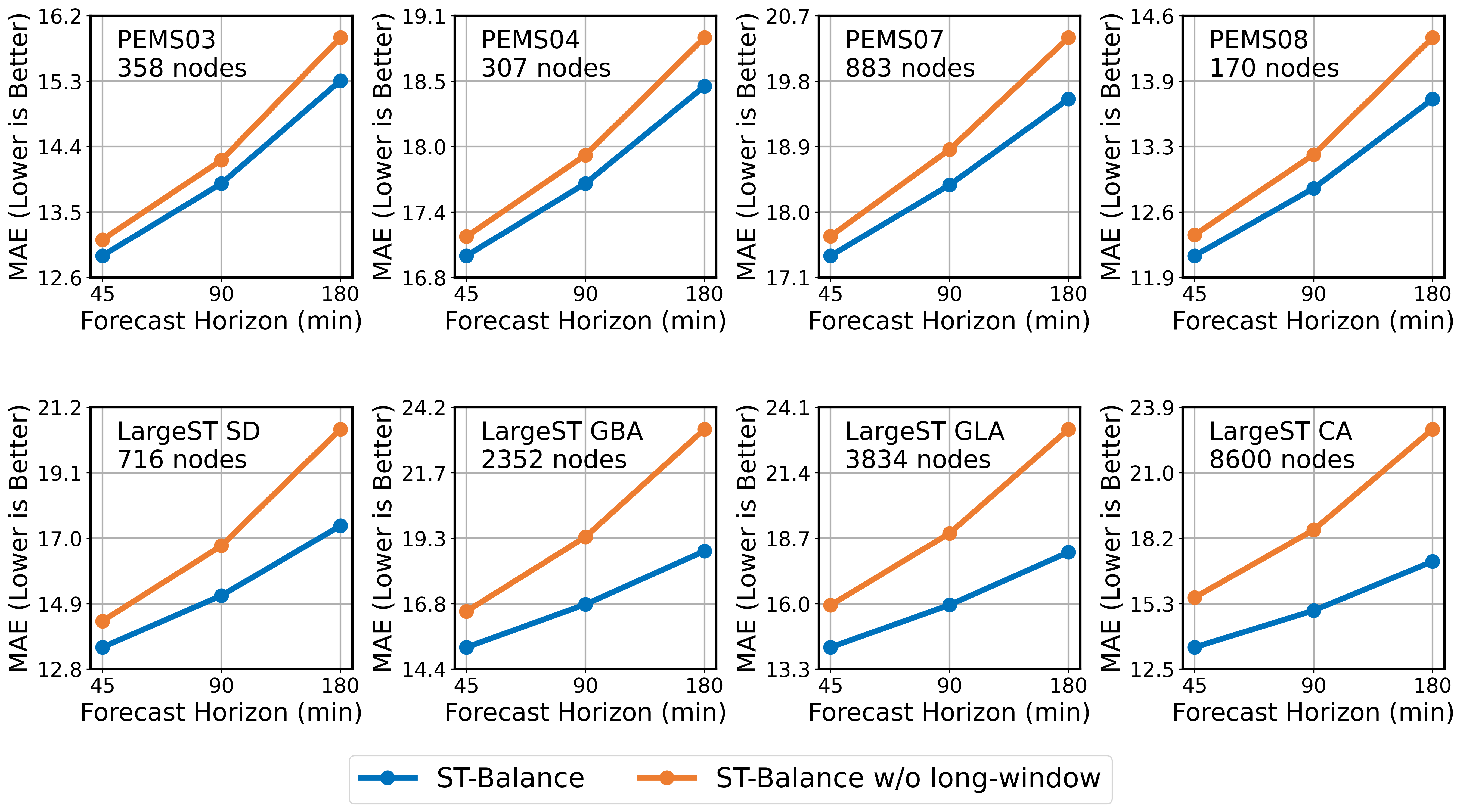}
	\end{center}
	\caption{\textbf{Impact of temporal module ablation on ST-Balance performance.} Prediction accuracy decreases notably without the long-window module, highlighting the module's critical contribution to capturing periodic traffic patterns.}
	\label{fig:without_T}
\end{figure*}

\begin{figure*}[htbp]
	\begin{center}
            \includegraphics[width=\linewidth]{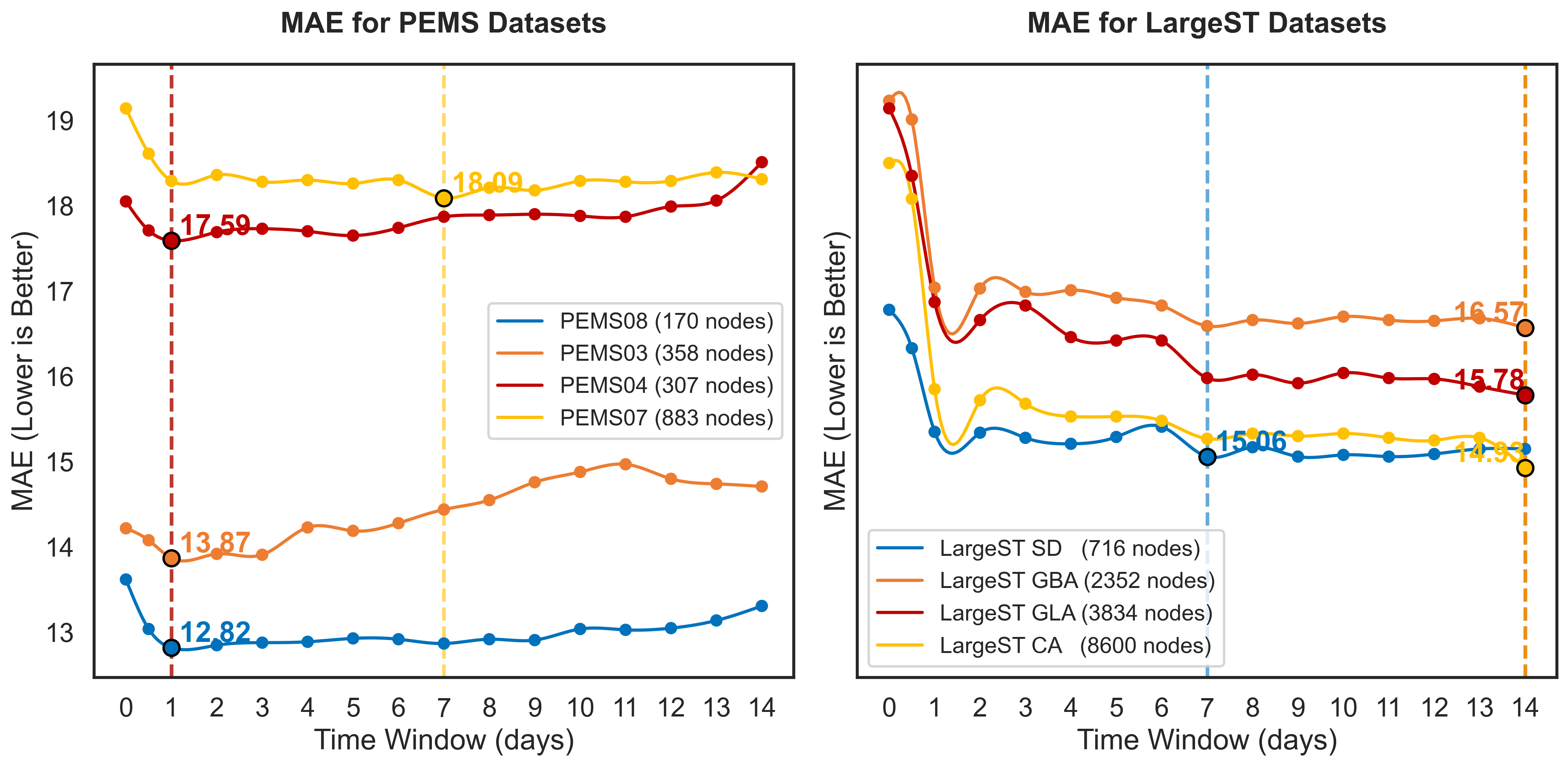}
	\end{center}
	\caption{\textbf{Sensitivity of ST-Balance performance to temporal window length.} MAE varies with increasing time window length, identifying optimal scales for different datasets.}
	\label{fig:long_window}
\end{figure*}
In purely temporal tasks, increasing the input sequence length can effectively capture extended temporal dependencies\cite{Autoformer, Informer2021}. However, in spatiotemporal settings, complex spatial–temporal interactions often make it computationally prohibitive to adopt similarly long sequences. As a result, many existing approaches constrain their analysis to short (e.g., 12-step) rolling windows, exemplified by configurations such as 5-minute horizons in PEMS and 15-minute horizons in LargeST. Even large pre-trained models (e.g., STEP\cite{STEP2022}, STD-MAE\cite{STD-MAE2024}) ultimately truncate longer sequences when integrated into spatiotemporal pipelines. Here, we investigate how extending the time window in ST-Balance confers substantial advantages, especially for large-scale datasets where intricate temporal patterns tend to emerge.

Figure \ref{fig:without_T} illustrates that the exclusion of the long-window module (w/o long-window) results in a notable degradation of ST-Balance's performance, particularly when applied to large-scale tasks and longer forecasting horizons. Conversely, the inclusion of an extended time window significantly improves the model's accuracy by effectively capturing daily and weekly traffic patterns. For example, as depicted in Figure \ref{fig:long_window}, a 24-hour window (288 steps) yields optimal performance for small-scale datasets (PEMS03, PEMS04, PEMS08). For medium-scale datasets, a 7-day window (2016 steps for PEMS07, 672 steps for LargeST SD) provides the best results, while a 14-day window (1344 steps) maximizes accuracy for large-scale datasets (LargeST GLA, GBA, CA).
However, excessively long windows can introduce redundancy or stale patterns, making temporal variability dominate and leading to diminishing returns in predictive accuracy. Notably, the empirically optimal window lengths tend to occur near the regime where temporal context becomes sufficient to constrain the effective spatial complexity after reduction; this regime corresponds to operating points closer to the diagonal in Figure~\ref{fig:Imbalance_Ratio}. This supports using entropy mismatch as a practical diagnostic for window selection rather than as a guarantee of monotonic improvement.

These findings indicate that temporal expansion, coupled with spatial dimensionality reduction, enables ST-Balance to process extended sequences without incurring prohibitive computational overhead. Aligning the time window with both the dataset’s intrinsic periodicities and its spatial scale allows ST-Balance to capture broader temporal dependencies and adapt to varying complexities. In practice, smaller datasets often benefit from shorter windows, whereas larger and sparser networks require extended windows to uncover long-range periodic patterns and more diffuse spatiotemporal correlations. This synergy, achieved by balancing spatial dimensionality reduction with long-range temporal signals, ultimately leads to superior spatiotemporal forecasting outcomes.

\subsection{Multi-domain applicability performance}
After confirming ST-Balance’s high accuracy in multi-scale, large-scale traffic forecasting, we next investigated its applicability capacity in the meteorological and public health domains. These evaluations serve to measure the model’s robustness in complex spatiotemporal scenarios characterized by markedly different data distributions, and highlight its potential for widespread real-world deployment.

\subsubsection{Meteorological Forecasting}
\begin{figure*}[htbp]
	\begin{center}
		\includegraphics[width=\linewidth]{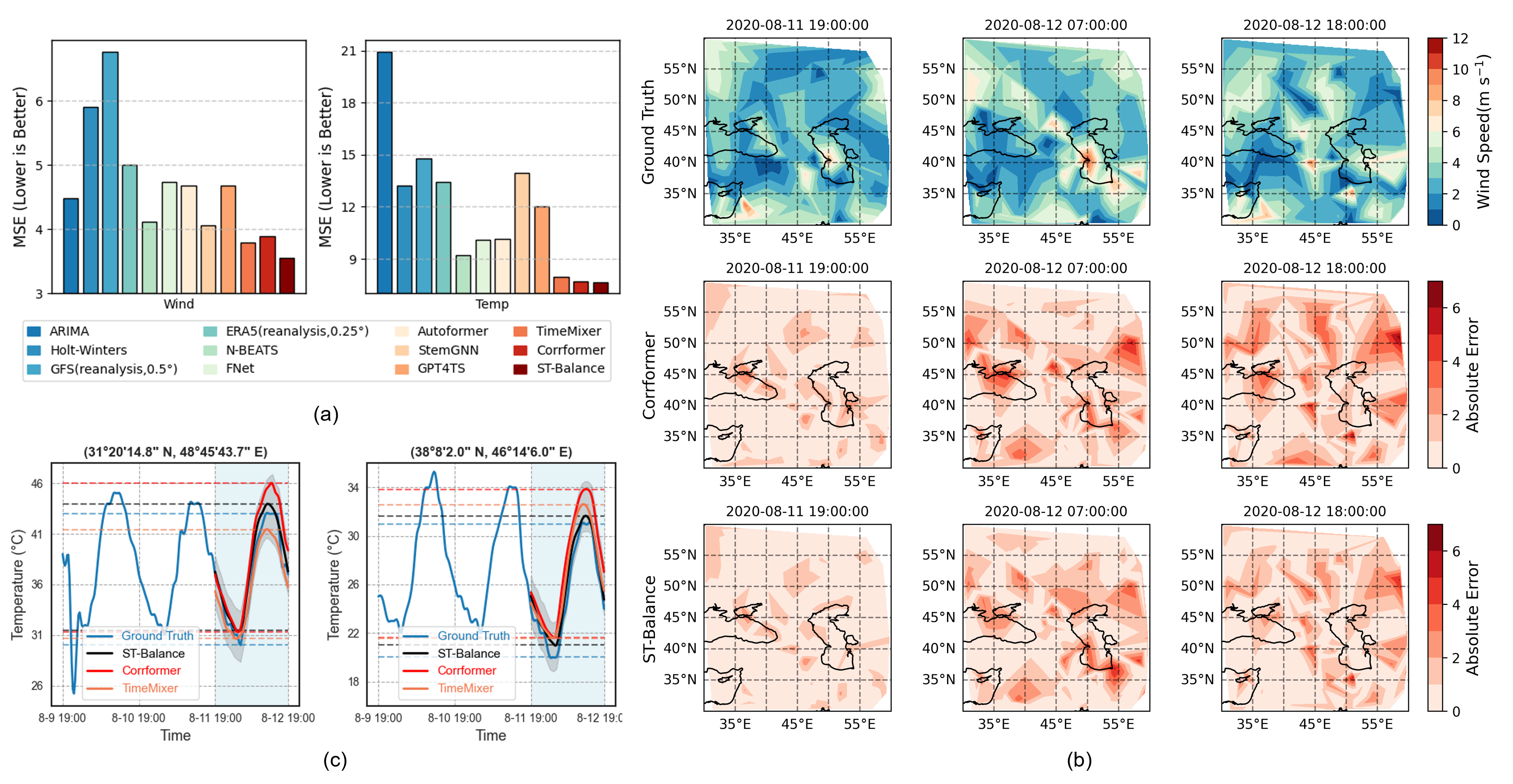}
	\end{center}
	\caption{\textbf{Model comparisons for meteorological forecasting.} (a) Aggregate performance (MSE) of ST-Balance versus benchmark models on Wind and Temp datasets. (b) Temporal evolution of prediction errors for wind speed forecasts, highlighting ST-Balance's reduced errors relative to Corrformer over extended periods. (c) Detailed temperature predictions at two spatial locations, emphasizing ST-Balance’s superior ability to model abrupt and irregular variations.}
	\label{fig:weather}
\end{figure*}
Meteorological prediction typically entails substantial spatial heterogeneity, nonlinear and non-stationary processes, and complex temporal dependencies. To systematically assess ST-Balance under these conditions, we employed two meteorological datasets, Wind and Temp. Figure \ref{fig:weather}(a) shows that ST-Balance achieves lower mean squared error (MSE) than statistical baselines (e.g., ARIMA\cite{ARIMA1976}, Holt–Winters \cite{Holt-Winters2018}), numerical meteorological systems (e.g., GFS\cite{GFS}, ERA5\cite{ERA5}), and various deep learning approaches (including N-BEATS\cite{N-BEATS2021}, FNet\cite{Fnet2021}, Autoformer\cite{Autoformer}, StemGNN\cite{StemGNN}, GPT4TS\cite{GPT4TS2023}, TimeMixer\cite{TimeMixer2024} and Corrformer\cite{Corrformer2023}). See Supplementary, 4.B 
for error metrics and Supplementary, 5.B 
for additional results.

These advantages are evident not only in aggregate metrics but also in localized analyses. Figure \ref{fig:weather}(b) depicts wind-speed forecasts at specific spatiotemporal coordinates, revealing that from 19:00 on 11 August to 18:00 on 12 August 2020, ST-Balance maintains lower errors over longer prediction horizons. Figure \ref{fig:weather}(c) focuses on temperature predictions at two spatial nodes across 72 hours, showing that ST-Balance not only captures diurnal temperature cycles but also adapts to abrupt, aperiodic shifts. By contrast, while Corrformer\cite{Corrformer2023} and TimeMixer\cite{TimeMixer2024} effectively traces the general trend, it exhibits slightly diminished accuracy in handling non-stationary, irregular fluctuations. Taken together, these results suggest that ST-Balance offers enhanced robustness and adaptability for modeling extended temporal dependencies and complex meteorological features.

\subsubsection{Epidemic Forecasting}
\begin{figure*}[htbp]
	\begin{center}
		\includegraphics[width=\linewidth]{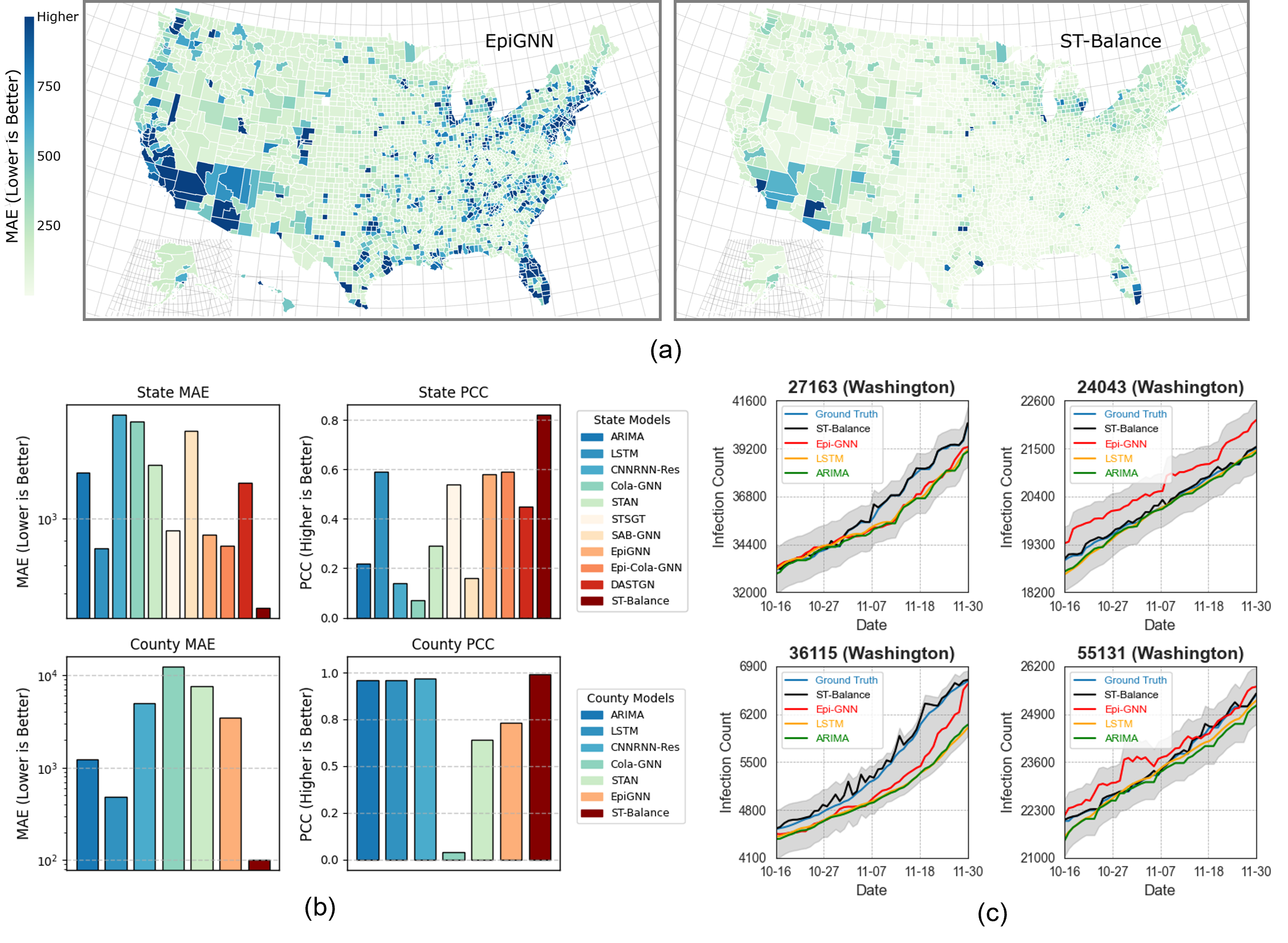}
	\end{center}
	\caption{\textbf{Comparative performance of different models in epidemic forecasting tasks.} (a) Average MAE of different models of Infection at the county-level. (b) Average infection MAE and PCC of different models on state and county level. (c) The forecasting local results of infections from Oct 16, 2021 to Nov 30, 2021.}
	\label{fig:epi}             
\end{figure*}

Epidemic data exhibit considerable spatiotemporal uncertainty and volatility, presenting significant challenges for predictive model applicability. ST-Balance, surpasses existing state-of-the-art methods at the state-level (51-node) predictions, including both infection (Figure \ref{fig:epi}(b)) and death count (Supplementary, 5.C 
), demonstrating notable accuracy across comparable scales.

To thoroughly evaluate the effectiveness of spatiotemporal balancing, we expanded the forecasting scope from the 51 state-level nodes to encompass 3,342 county-level nodes. Under this extensive and detailed scale, the superior predictive capacity of ST-Balance becomes increasingly evident. An intriguing observation emerges at this spatial granularity: the predictive performance of traditional graph neural network (GNN)-based methods significantly deteriorates as spatial complexity increases. Several GNN-based approaches either encounter memory constraints (e.g., STSGT\cite{STSGT2022}, SAB-GNN\cite{SAB-GNN2022}, DASTGN\cite{DASTGN2024} ) or demonstrate inferior forecasting performance (e.g., Cola-GNN\cite{Cola-GNN2020}, CNNRNN-Res\cite{CNNRNN-Res2018}, STAN\cite{STAN2021}, EpiGNN\cite{EpiGNN2022}) compared to simpler baseline methods ARIMA\cite{ARIMA1976} and LSTM\cite{LSTM}, highlighting inherent limitations in generalizing across spatial scales.

In contrast, ST-Balance consistently demonstrates superior predictive performance, achieving an MAE of 100.26 (Figure \ref{fig:epi}(b)), significantly outperforming competing methods in forecasting infection dynamics. This performance enhancement underscores the model's robustness in addressing increased spatial granularity. Visualized predictions and comparative analyses at both state and county levels are provided in Figure \ref{fig:epi}(a), with further detailed analyses available in Supplementary, 5.C 
.

Moreover, a localized assessment was conducted at the county scale to investigate model consistency. Four counties named Washington were randomly selected for this analysis, as depicted in Figure \ref{fig:epi}(c). This localized evaluation highlights ST-Balance's consistency and adaptability in capturing intricate spatiotemporal epidemic patterns across geographically diverse counties.

\section{Discussion}
We propose ST-Balance, a scalable framework for spatiotemporal forecasting that mitigates spatial--temporal mismatch by (i) compressing spatial structure with low-rank embeddings and (ii) expanding usable temporal context via a multi-scale temporal enhancement module. Extensive experiments on traffic, meteorological, and epidemic datasets demonstrate consistent accuracy improvements across spatial scales and forecasting horizons, while keeping memory and runtime feasible on large graphs.

Our contribution is not centered on proposing a new backbone, but on a practical and reusable design strategy for large-scale spatiotemporal modeling. The proposed spatial reduction and temporal expansion are plug-and-play components that can be integrated into strong existing predictors to further reduce error, with benefits becoming more pronounced as spatial scale increases. This suggests that improving the performance frontier of current methods does not necessarily require introducing a wholly new architecture, but can be achieved by rebalancing effective spatial complexity and temporal context under limited capacity.

We use spatial and temporal entropy as practical diagnostics to guide design choices, rather than as a proof that aligning \(H_S\) and \(H_T\) guarantees optimal forecasting. The optimal operating point depends on model capacity, regularization, and data non-stationarity; overly long windows may introduce redundancy or stale patterns. Future work will further characterize robustness when spatial dependency is weak, and explore drift-aware mechanisms for selecting or weighting long-range history, as well as stronger spatial compression and adaptive clustering guided by local connectivity.

\bibliographystyle{elsarticle-num}
\bibliography{sn-bibliography}

\end{CJK*}
\end{document}